\documentclass[conference]{IEEEtran}
\IEEEoverridecommandlockouts
\usepackage{cite}
\usepackage{amsmath,amssymb,amsfonts}
\usepackage{algorithmic}
\usepackage{graphicx}
\usepackage[outdir=./]{epstopdf}
\usepackage{textcomp}
\usepackage{xcolor}
\def\BibTeX{{\rm B\kern-.05em{\sc i\kern-.025em b}\kern-.08em
    T\kern-.1667em\lower.7ex\hbox{E}\kern-.125emX}}
    
\makeatletter

\def\ps@IEEEtitlepagestyle{%
  \def\@oddfoot{\mycopyrightnotice}%
  \def\@evenfoot{}%
}
\def\mycopyrightnotice{%
  {\footnotesize 978-0-7381-2333-2/20/\$31.00~\copyright~2020 IEEE\hfill}
  \gdef\mycopyrightnotice{}
}

\@ifundefined{showcaptionsetup}{}{
 \PassOptionsToPackage{caption=false}{subfig}}
\usepackage{subfig}
\makeatother

\usepackage{eso-pic}
\newcommand\AtPageUpperMyright[1]{\AtPageUpperLeft{%
 \put(\LenToUnit{0.5\paperwidth},\LenToUnit{-1cm}){%
     \parbox{0.5\textwidth}{\raggedleft\fontsize{9}{11}\selectfont #1}}%
 }}%
\newcommand{\conf}[1]{%
\AddToShipoutPictureBG*{%
\AtPageUpperMyright{#1}
}
}

\begin{document}

\title{Restyling Images with the Bangladeshi Paintings Using Neural Style Transfer: A Comprehensive Experiment, Evaluation, and Human Perspective}

\conf{2020 23\textsuperscript{rd} International Conference on Computer and Information Technology (ICCIT), 19-21 December, 2020.} 

\author{\IEEEauthorblockN{Manal}
\IEEEauthorblockA{\textit{Computer Science and Engineering} \\
Shahjalal University of Science\\ and Technology\\
Sylhet, Bangladesh\\
aymaanmanal@gmail.com}
\and
\IEEEauthorblockN{Ali Hasan Md. Linkon}
\IEEEauthorblockA{\textit{Computer Science and Engineering} \\
Shahjalal University of Science\\ and Technology\\
Sylhet, Bangladesh\\
linkon3.1416@gmail.com}
\and

\IEEEauthorblockN{Md. Mahir Labib}
\IEEEauthorblockA{\textit{Computer Science and Engineering} \\
Shahjalal University of Science\\ and Technology\\
Sylhet, Bangladesh\\
mdmahirlabib@gmail.com}
\and
\IEEEauthorblockN{Marium-E-Jannat}
\IEEEauthorblockA{\textit{Computer Science and Engineering} \\
Shahjalal University of Science and Technology\\
Sylhet, Bangladesh\\
jannat-cse@sust.edu}
\and 

\IEEEauthorblockN{Md Saiful Islam}
\IEEEauthorblockA{\textit{Computer Science and Engineering} \\
Shahjalal University of Science and Technology\\
Sylhet, Bangladesh\\
saiful-cse@sust.edu}

}

\maketitle

\begin{abstract}
In today’s world, Neural Style Transfer (NST) has become a trendsetting term. NST combines two pictures, a content picture and a reference image in style (such as the work of a renowned painter) in a way that makes the output image look like an image of the material, but rendered with the form of a reference picture. However, there is no study using the artwork or painting of Bangladeshi painters.  Bangladeshi painting has a long history of more than two thousand years and is still being practiced by Bangladeshi painters. This study generates NST stylized image on Bangladeshi paintings and analyzes the human point of view regarding the aesthetic preference of NST on Bangladeshi paintings. To assure our study's acceptance, we performed qualitative human evaluations on generated stylized images by 60 individual humans of different age and gender groups. We have explained how NST works for Bangladeshi paintings and assess NST algorithms, both qualitatively \& quantitatively.  Our study acts as a pre-requisite for the impact of NST stylized image using Bangladeshi paintings on mobile UI/GUI and material translation from the human perspective. We hope that this study will encourage new collaborations to create more NST related studies and expand the use of Bangladeshi artworks.
\end{abstract}

\section{Introduction}
Artworks create emotional responses among people, and it has a far more significant social, economic, and geopolitical impact on our life. Artworks enlighten the human mind and help in understanding reality and imagination at the same time. Bangladeshi painting is a visual arts style in which painters portray Bangladesh. Every artist has their distinct painting styles and strokes. Regenerating any painting requires much time and critical skill. 

In the last decade, machine learning and deep learning technology have improved significantly. Deep learning can learn high-level features from the image. Moreover, with the help of massively parallel computing (GPUs) in recent years, deep learning techniques have achieved immense popularity in style transfer. Neural Style Transfer(NST) is used to transfer an artist's style to another particular image, producing a new art piece. Using these stylized images, we can have a presence of mind of the artists. 

Earlier, researchers propose various style transfer strategies. In the paper of Gatys et al.\cite{Gatys}, motivated by the influence of the Convolutional Neural Network (CNN), they implemented the use of CNN as a replication on original images of many renowned art styles. Their algorithm's key idea is to refine an image iteratively to match the ideal CNN function distribution, including both pieces of knowledge about photo quality and painting style. Gatys et al. \cite{Gatys} opened a new era in the field of style transfer. NST is a technique used for optimization to take two images - a content image and a style image. Higher-level features like the image's content extract from the content image, and low-level features like colors, textures extract from the style image. This blended image generates a new image containing the color, texture from the style image, and contents from the content image. Over the last few years, NST has proven to be an exciting area of study inspired by technological problems and industry demands.
\begin{figure}[htbp]
\centerline{\includegraphics[scale=0.11]{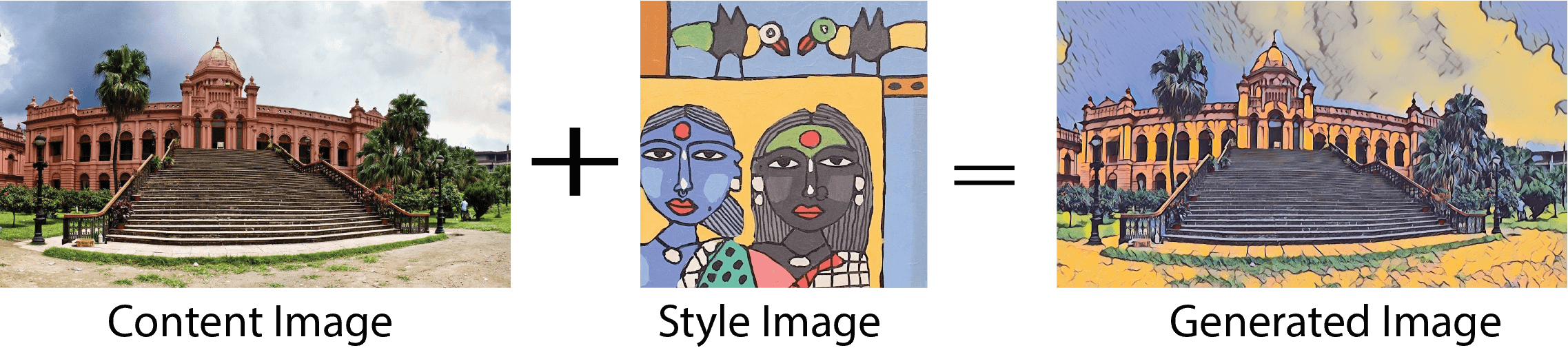}}
\caption{Generated image from the content and style image. The generated image has the image contents from the content image \& color, texture from the style image.}
\label{aestheticPreference}
\end{figure}

There are three things that a style transfer model needs:
\begin{itemize}
\item  \textbf{Generating model:} It would generate the output images. 
\item  \textbf{Loss functions:} Good choice of loss functions is essential to achieve good results.
\item  \textbf{Loss Network:}  A pre-trained CNN model can extract good features from the images. 
\end{itemize}

NST based photo filter applications, like Prisma, have received immense popularity among users in recent days. Social media apps like Snapchat, Instagram also use NST based filters for aesthetic image modification.
Fischer et al.\cite{Fischer} aim to create a system that enables content creators or consumers to redesign their favorite app interface and effectively bring to renowned designers' talents at their fingertips. They try to shift the NST trend towards GUI/UI. They anticipate a world in which users can expect to see beautiful designs in any application they are using and experience a wide range of designs as they do in fashion today. In the paper of Garcia et al.\cite{Garcia}, they suggest enhancing CNN-based content translation functionality, for instance, normalization whitening using a VGG19\cite{VGG} network. An efficient image recall process is used to find an optimal image that better translates object regions. To convert regions into target domains, they can synthesize images using the traditional NST process paired with a simultaneous real-time segmentation technique. Fortunately, these works using diverse art styles allow rising aesthetic pleasure among consumers associated with industrial or social media platforms.

As we see, there are many applications regarding NST in modern days, but there is no study on Bangladeshi painting related to this area. We discuss the latest progress of NST in this paper using Bangladeshi artworks. Every painter in Bangladesh has a unique style and medium. So their painting style is completely different from each other. We have used various Bangladeshi artists' paintings and tried to re-generate new images using their paintings style. We have presented qualitatively and quantitatively two types of assessment approaches on our generated images. We have analyzed if people can differentiate between the original painting and stylized painting generated using NST. To test our generated images' quality and aesthetic preference, we have conducted an acceptance analysis of 60 university students in Bangladesh. The analysis ends with a debate on various NST applications such as mobile UI/GUI on Bangladeshi paintings and open questions for study in the future.

\section{Related Work}
Gatys et al. \cite{Gatys} have contributed significantly to the early stage of Neural Style Transfer. This paper used image optimization-based NST, where CNN image representations were designed for object recognition, rendering high-level image information visible. Gatys et al. 's algorithm\cite{Gatys} allowed the development of new high-perceptive images combining the contents of an arbitrary picture with the appearance of various famous works of art. This transfer form algorithm combined a parametrically based texture model with an image reversal process. Using state-of-the-art convolutional neural networks feature representations, this NST method was a texture transfer algorithm that restricts a texture synthesis process. The style transfer process was time-consuming and procedural, but in comparison, the visual quality results were attractive. It was, therefore, generally considered the fundamental approach for modern NST.

Every literature based on Per-Style-Per-Model only worked in one style. Ulyanov et al.\cite{Ulyanov} and Johnson et al.\cite{Johnson} stylized findings were slightly identical. It was due to their identical concept and varied only in their comprehensive network architectures.  The figures were somewhat less promising for Li and Wand\cite{LiWand}. Since it was focused on the Generative Adversarial Network (GAN)\cite{GAN}, the training cycle wasn't that accurate to some extent.

Johnson et al.\cite{Johnson} proposed perceptual loss functions in image transformation work for feed-forward network training. They showed image-style transmission results of Gatys et al. \cite{Gatys} 's optimization problem in real-time. They also experimented with a one-image super-resolution, which replaced a pixel loss with a perceptive loss, visually delighted results. They showed that perceptual loss training allows for better reconstruction of fine details and edges.

Ulyanov et al. \cite{Ulyanov} proposed an approach that moves the computational burden to a learning level. The method formed compact feed convolutional neural networks, generates multiple samples of the same arbitrary texture, and transfers the artistic style from a given image to any other image. Their result was 100 times faster than Gatys et al. \cite{Gatys}. They could produce complex texture for complex loss functions. Typically their process yielded excellent synthesis of the texture.

Li and Wand et al. \cite{LiWand} have proposed the Markovian Generative Adversarial Networks (MGANs), a method for training generative neural networks for efficient texture synthesis. Their strategy aims at making deep Markovian texture synthesis more effective. Their methodology sought to enhance deep Markov texture synthesis performance. This model was able to translate brown noise into natural texture and pictures into a form of art. The solution is 500 times quicker than  Gatys et al. \cite{Gatys}.

Multiple styles per model are another example of the style transfer process.  In this method, various styles are used in a single model. The algorithm Dumoulin et al. \cite{Dumoulin} and Chen et al. \cite{Chen} developed gives a solution to connect a limited number of parameters for each style. Both designed the same architecture and resulted in their algorithms. Although the results of Dumoulin et al. and Chen et al. 's algorithms were promising, their model size expanded as the types of learning increased.

The Huang and Belongie \cite{Huang} algorithms are designed to successfully suit global summary feature statistics and boost visual quality. They used the layer of adaptive normalization instance (AdaIN), which combines the mean and the variance of the contents with the features of form. Their method achieves the speed of the fastest existing solution, without being limited to any kinds. Their approach also allows user-adaptable controls, including interpolation in content type, interpolation in form, spatial color constraints with a single neural feed-forward network. However, their algorithm does not seem successful in controlling intricate patterns, and their level of styling is still linked to the variety of training. AdaIN passes the theme of the feature Space transfers features statistics, particularly the mean and variance of the channel.

Fischer et al.\cite{Fischer} aims to create a framework that allows content creators or users to reinvent an app interface of their preference and effectively bring renowned designers' talents to their fingertips. They are the first to try moving the style to GUI. They envision a future in which users would expect to see beautiful designs in each application they use and experience a wide variety of designs as they can today in fashion. They demonstrate how ImagineNet can be used to redefine a range of applications, from the use of a static collection of assets to those integrating content and those creating an on-the-fly graphical user interface. The effects of the transition of the paintings' style to some GUI were seen in this paper. As the structural loss term is measured across layers, ImagineNet can not be used to such optimization techniques immediately. The algorithm can not be run on cell phones in real-time. More work is needed in the future to optimize ImagineNet.

\section{Dataset \& Image Source}
As there is no dataset of Bangladeshi paintings,  we have used open-source artworks of different artists that are allowed to use for educational purposes as our style images. Most of the paintings are used from WikiArt\cite{WikiArt}, an online, user-editable visual art encyclopedia. We have selected famous artworks of renowned painters of Bangladesh, such as SM Sultan, Abdus Shakoor, Hashem Khan, Qayyum Chowdhury, etc. Content images used in this paper are also open-sourced and fair use. To train the model architecture explained in the paper of Johnson et al.\cite{Johnson}, we have used the test images from MS-COCO dataset\cite{MSCOCO} to pre-train each style image. There are a total of 40775 images in the MS-COCO test dataset. We train the style model using the whole dataset.

\section{Methodology}
From the different NST approach, we have selected Gatys et al.\cite{Gatys} and Johnson et al.\cite{Johnson} for our study. Both of them are computationally efficient and generate excellent stylized images. We generate stylized images using these two methods on Bangladeshi paintings.

\subsection{Data Preprocessing}
In this paper, we have used diverse styles and content images. Firstly, we converted our style and content images to $512\times512$ pixels for the architecture proposed in the paper of Gatys et al.\cite{Gatys}. To run the architecture introduced in the paper of Johnson et al.\cite{Johnson}, we converted our images to $256\times256$ pixels. 

\subsection{Proposed Techniques}
In this paper, we have tried to achieve realism in our stylized images prepared using the architectures of Gatys et al.\cite{Gatys} and Johnson et al.\cite{Johnson}. Gatys et al.\cite{Gatys} are considered as the gold standard for NST. Johnson et al.\cite{Johnson} provide an offline service through building an offline model that can be used multiple times to generate stylized images within a few seconds. We have tested the images using external participants. Based on the participants' point of view, we analyzed and evaluated the paintings. From the consensus, it is proved that NST produces a magnificent result that confuses if the stylized images are real artwork by an artist or generated using deep learning techniques.

\subsection{Human Evaluations}
We have evaluated our stylized images by taking opinions from participants. We have randomly selected a group of 60 students from the different universities of Bangladesh. Among 60 participants, there are 20 female and 40 male observations within the range of age 20-30 years. All of the 60 participants are undergrad and graduate students—furthermore, most of them have decent knowledge regarding painting and style transfer. We have collected our data using Google Forms. We asked them whether the stylized images looked like real painting or not. The result is measured in terms of binary recognition of an image, whether it looks like an original painting or not. We have tested the stylized image generated from the architectures of both Gatys et al.\cite{Gatys} and Johnson et al.\cite{Johnson}. The observers rated the aesthetic preference of the stylized images on a scale of 1 to 5. We have also collected participants' opinions regarding stylized GUI \& UI and their viewpoints concerning NST.

\section{Hardware and Software}
We have used Nvidia Tesla K80 12GB GPU provided by Google Colaboratory. All the codes are implemented using PyTorch\cite{PyTorch} deep learning library. We have used both Python 2 and Python 3 language as our requirements.

\section{Analysis}

\subsection{Quantitative Analysis}

\begin{figure*}[htbp]
\centerline{\includegraphics[scale=0.48]{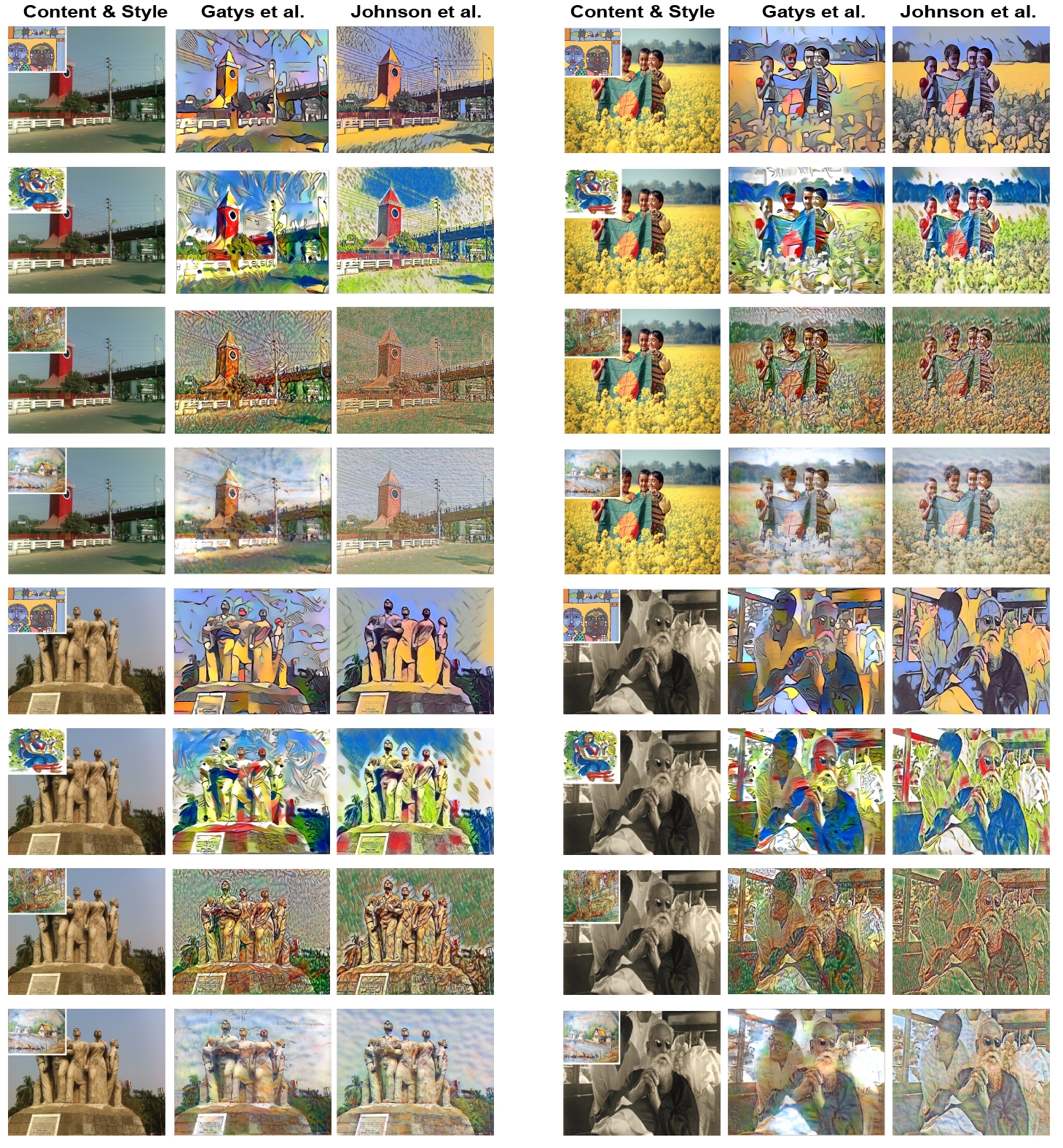}}
\caption{Generated stylized image using Gatys et al.\cite{Gatys} and Johnson et al.\cite{Johnson}}
\label{Combined}
\end{figure*}

\subsubsection{Running Time}
It took 150.3 seconds to run the architecture of Gatys et al.\cite{Gatys} for 500 iterations with an image size of $512\times$512. If an image size of $256\times$256 is taken, the computation time takes nearly as half as before. However, in the case of Johnson et al., 500 iterations with batch size four and an input image size of $256\times$256 took nearly 146.7 seconds to run. If an image size of $512\times$512 is taken, the computation time doubles. If the batch size is set to 1, it takes less time than the architecture implemented by Gatys et al.\cite{Gatys} by a few seconds.

\subsection{Qualitative Analysis}
Qualitative analysis is a complicated task for any stylized image. The message or representation of an image of any painter varies from person to person. Also, texture designs, tools, and mediums used to make an immense difference. Consequently, determining the aesthetic criteria for a stylized artwork is not easy. Different people have varying or contradictory views of the same stylized result. Example painting stylized results are shown in Figure \ref{Combined}.

To evaluate the output image's aesthetic scores, we have collected 60 participants' data using different NST algorithms given the same style and content image.  We asked for a scale of 1 to 5  aesthetic preference scores for each deep learning generated stylized images. We have seen that images are generated from Johnson et al.\cite{Johnson}  are mostly favored by the participants. All of the participants' aesthetic preference scores are represented in Figure \ref{aestheticPreference}.

\begin{figure}[htbp]
\centerline{\includegraphics[scale=0.4]{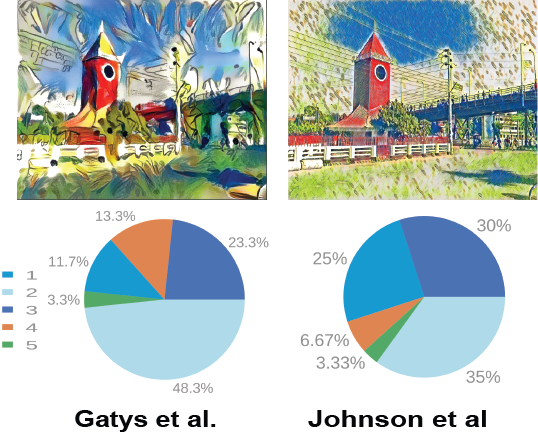}}
\caption{Aesthetic preference scores for the outputs of Gatys et al.\cite{Gatys} and Johnson et al.\cite{Johnson}. Aesthetic preference scores scale of 1 to 5 (1 represents low, 5 represents high)}
\label{aestheticPreference}
\end{figure}

We have also tried to determine whether the participants can differentiate any deep learning generated image from the original Bangladeshi painting. On average, 32\% of participants could not differentiate the stylized image. Even 51.7\% of participants recognize a painters' painting as an NST generated image. Figure \ref{PainterWhich} represents that deep learning models should improve more to generate a top-level of paintings.  \\

\begin{figure}[htbp]
\centerline{\includegraphics[scale=0.4]{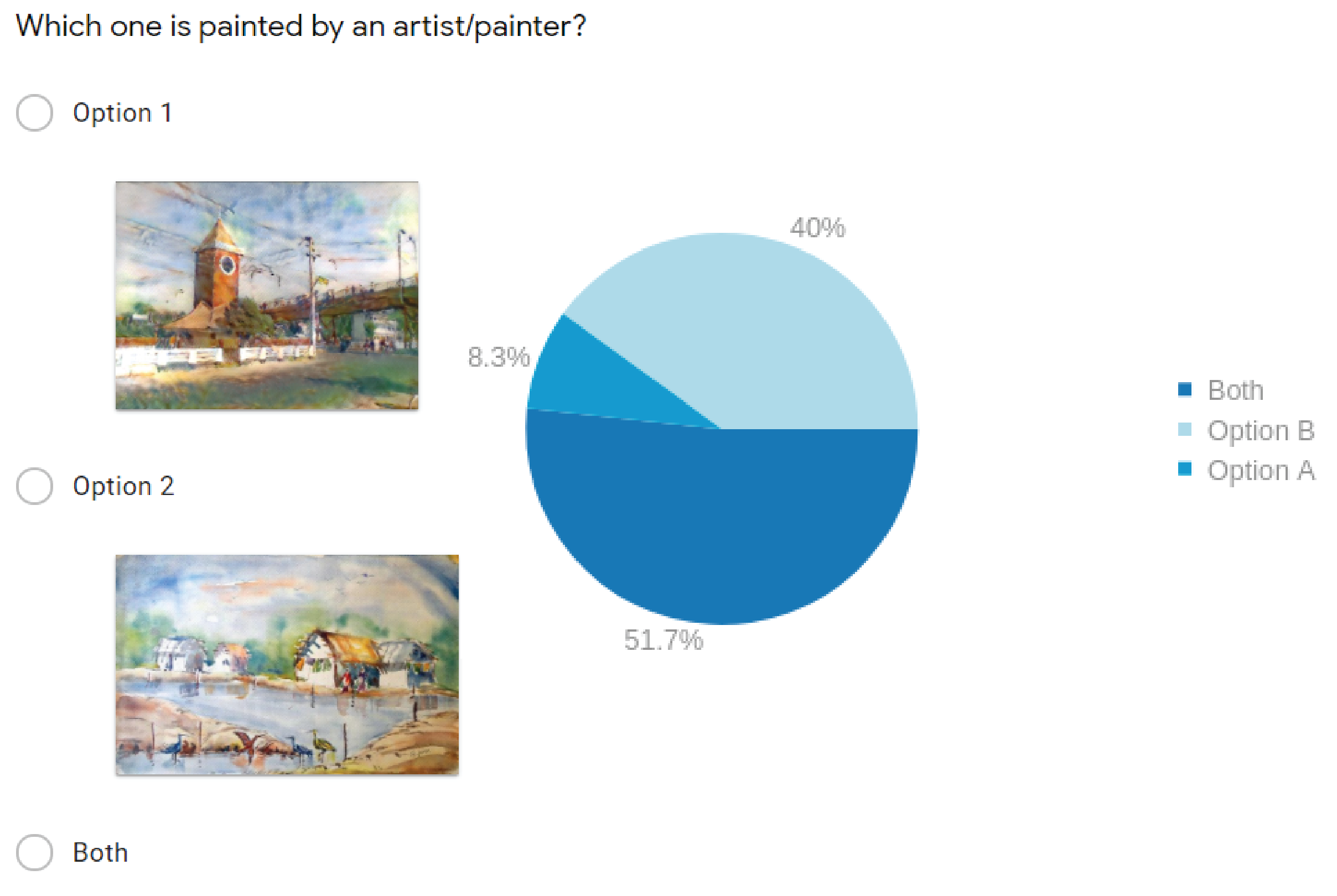}}
\caption{Option A image is generated by our deep learning model, and Option B image is the artwork of famous Bangladeshi painter Hashem Khan. But only 40\% of participants could recognize the original painting.}
\label{PainterWhich}
\end{figure}

\begin{figure}[htbp]
\centerline{\includegraphics[scale=0.45]{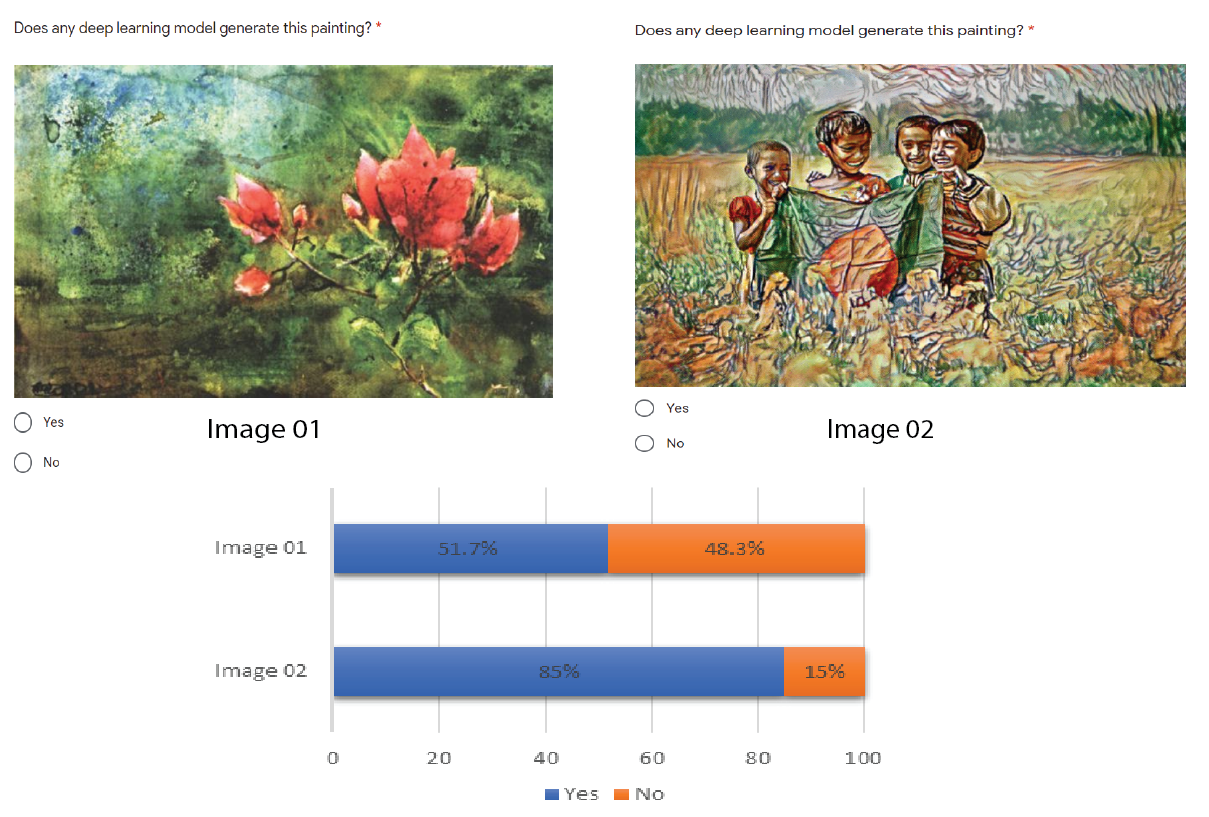}}
\caption{Image 01 is a real painting and Image 02 is generated by NST}
\label{PainterWhich}
\end{figure}

\begin{figure}[htbp]
\centerline{\includegraphics[scale=0.42]{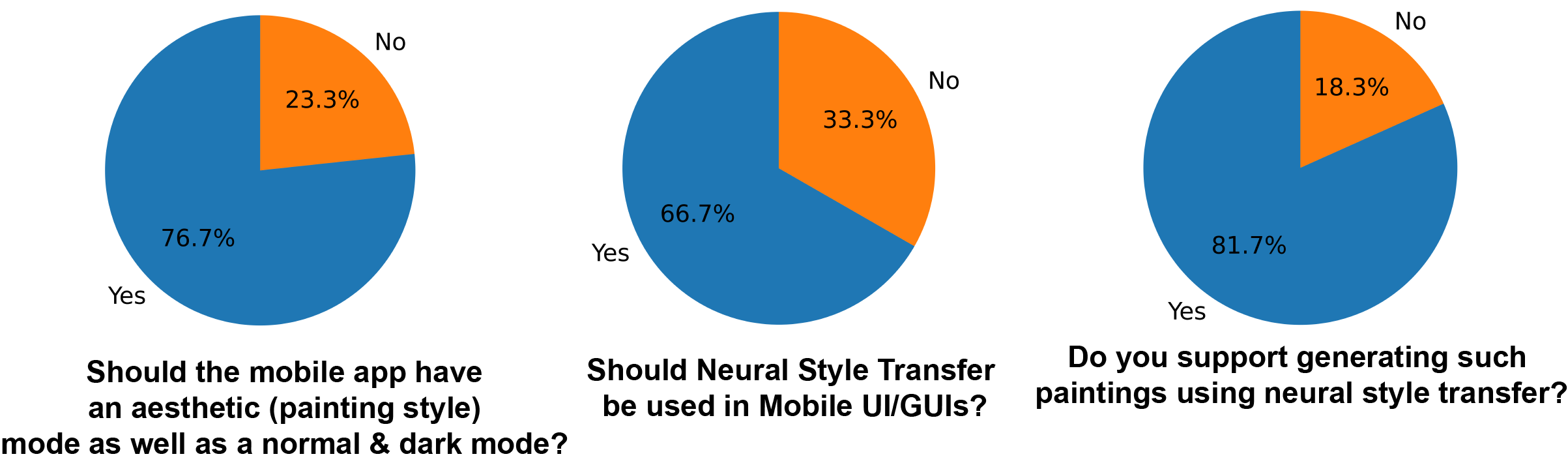}}
\caption{Views of participants regarding NST and  NST stylized mobile UI}
\label{Opinion}
\end{figure}

In Fischer et al.\cite{Fischer}, they re-styling apps GUI using neural style transfer. So we have tried to re-style several application GUI/UI using Bangladeshi painters' painting. 66.7\% of our participants highly appreciate this kind of aesthetic mobile GUI/UI. Furthermore, 76.7\% of our participants think mobile applications should have aesthetic mode besides dark and normal mode. As we have performed this research on a small scale of 60 participants, this study can be done on a large scale for better results. NST based mobile UI/GUI is an excellent source of future research. 81.7\% of our participants appreciate image style transfer and generate paintings using NST. Figure \ref{Opinion} represents the survey results of participants' views regarding NST and  NST stylized mobile UI.

\section{Discussion and Future Directions}
Paintings have a significant impact on our life. Different artists have different expression style. Colorful paintings are more pleasant to the eyes and thereby produce aesthetically pleasing stylized images. If aesthetically pleasing paintings are used to stylize different photographs of natural scenarios, it provides a realistic experience to the users. This paper has divided our works into two sections \\
• Running architectures of the two most famous Neural Style Transfer paper on Bangladeshi paintings and comparing their performance  \\
• Analysis of data collected from the consensus on the stylized images being real or not. 

From our experiments and data collection, We have seen that Gatys et al.\cite{Gatys} and Johnson et al.\cite{Johnson} both produce the best quality stylized images, but participants prefer Johnson et al.\cite{Johnson} stylized images on Bangladeshi paintings.  In the future, paintings can be used to neural style transfer in different materials and objects in our day to day use\cite{Garcia}. As a result, an artist's value can be preserved and can regenerate their works easily using deep learning technology. 

In the future, NST in deep learning can create an overall impression among people. NST will positively influence mobile and other technology GUI and UI\cite{Fischer}. We have planned to integrate Bangladeshi paintings NST based mobile app UI \& we are currently working on this. There is a lack of open-source \& high-quality images of paintings of different Bangladeshi painters. A collection of style images of different Bangladeshi painters is necessary for carrying out broad-scale research in this field.  We hope more enthusiastic contributors will come forward and contribute significantly to Bangladeshi painting stylized images.

\section{Conclusion}

In recent years, NST has become an exciting area of study for academic and industry sectors. We tried some NST algorithm on Bangladeshi paintings and showed the result. We have evaluated the algorithms quantitatively \& qualitatively. We also take a survey of people about the NST generated images and even take a peer analysis of the images generated by the NST to see if they can distinguish between the original one and the one generated.  For our qualitative study, we analyzed their feedback to the generated image. Since there is no such work using Bangladeshi artwork, it could be the guideline for NST on painting in Bangladesh.

\section*{Acknowledgment}
We would like to thank Shahjalal University of Science and Technology (SUST) and SUST NLP research group for their support.

\end{document}